\documentclass[letterpaper]{article} 
\usepackage{arxiv}  
\usepackage{times}  
\usepackage{helvet}  
\usepackage{courier}  
\usepackage[hyphens]{url}  
\usepackage{graphicx} 
\usepackage{natbib}  
\usepackage{caption} 
\usepackage{algorithm}
\usepackage{algorithmic}
\usepackage[colorlinks=true, linkcolor=black, citecolor=black, urlcolor=black]{hyperref}

\usepackage{newfloat}
\usepackage{listings}
\DeclareCaptionStyle{ruled}{labelfont=normalfont,labelsep=colon,strut=off} 
\floatstyle{ruled}
\newfloat{listing}{tb}{lst}{}
\floatname{listing}{Listing}
\title{Seeing the Unseen: Zooming in the Dark with Event Cameras}
\author{
    Dachun Kai\textsuperscript{\rm 1},
    Zeyu Xiao\textsuperscript{\rm 2},
    Huyue Zhu\textsuperscript{\rm 1},
    Jiaxiao Wang\textsuperscript{\rm 1},
    Yueyi Zhang\textsuperscript{\rm 3},
    Xiaoyan Sun\textsuperscript{\rm 1,4}\thanks{Corresponding author.}
}
\affiliations{
    \textsuperscript{\rm 1}MoE Key Laboratory of Brain-inspired Intelligent Perception and Cognition, University of Science and Technology of China\\
    \textsuperscript{\rm 2}National University of Singapore \\
    \textsuperscript{\rm 3}Miromind AI\\
    \textsuperscript{\rm 4}Institute of Artificial Intelligence, Hefei Comprehensive National Science Center\\
    dachunkai@mail.ustc.edu.cn, sunxiaoyan@ustc.edu.cn
}

\usepackage{bibentry}
\usepackage{amssymb}
\usepackage{amsmath}
\usepackage{multirow} 
\usepackage{xcolor}   
\usepackage{booktabs} 
\usepackage{colortbl} 
\usepackage{makecell}

\newcommand{\ie}{\textit{i.e.}}

\begin{document}

\maketitle

\begin{abstract}
This paper addresses low-light video super-resolution (LVSR), aiming to restore high-resolution videos from low-light, low-resolution (LR) inputs. Existing LVSR methods often struggle to recover fine details due to limited contrast and insufficient high-frequency information. To overcome these challenges, we present RetinexEVSR, the first event-driven LVSR framework that leverages high-contrast event signals and Retinex-inspired priors to enhance video quality under low-light scenarios. Unlike previous approaches that directly fuse degraded signals, RetinexEVSR introduces a novel bidirectional cross-modal fusion strategy to extract and integrate meaningful cues from noisy event data and degraded RGB frames. Specifically, an illumination-guided event enhancement module is designed to progressively refine event features using illumination maps derived from the Retinex model, thereby suppressing low-light artifacts while preserving high-contrast details. Furthermore, we propose an event-guided reflectance enhancement module that utilizes the enhanced event features to dynamically recover reflectance details via a multi-scale fusion mechanism. Experimental results show that our RetinexEVSR achieves state-of-the-art performance on three datasets. Notably, on the SDSD benchmark, our method can get up to 2.95 dB gain while reducing runtime by 65\% compared to prior event-based methods.
\end{abstract}

\begin{links}
    \link{Code}{https://github.com/DachunKai/RetinexEVSR}
\end{links}

\section{Introduction}

    Video super-resolution (VSR) aims to restore high-resolution (HR) videos from low-resolution (LR) inputs. While existing methods~\cite{zhou2024video} get good results on general videos, they often fail under low-light conditions. However, such conditions are common in real-world applications, such as video surveillance, where zooming in on distant license plates or human faces at night is often required. Other important scenarios include remote sensing~\cite{xiao2025multi} and night videography~\cite{yue2024unveiling,li2025difiisr}. Therefore, it is essential to develop VSR algorithms specifically designed for low-light videos.

    \begin{figure}[t!]
        \centering
        \includegraphics[width=\columnwidth]{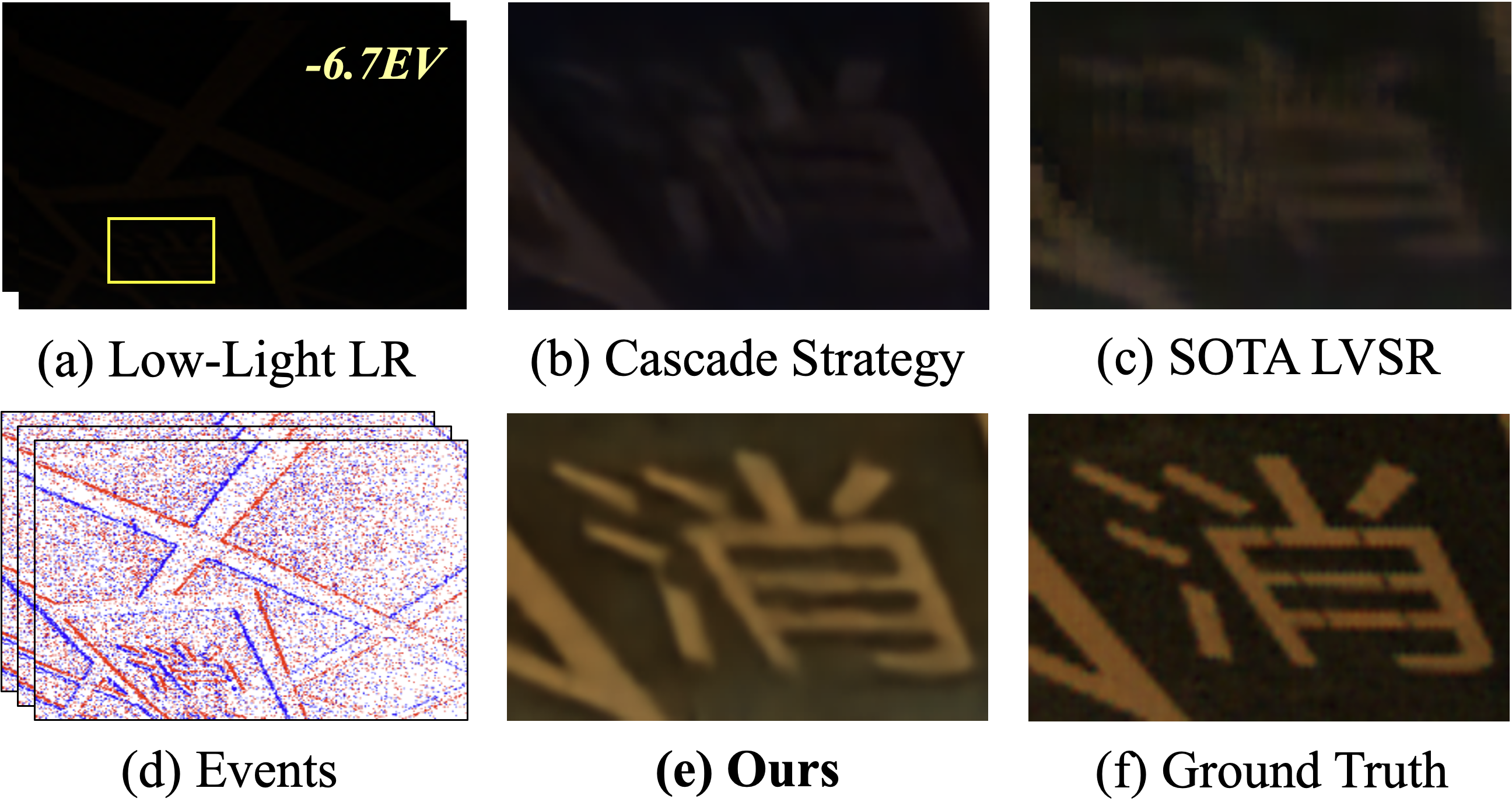}
        \caption{An example (a) from an extremely low-light (-6.7 EV) LR sample, enhanced by (b) SOTA LVE~\cite{li2023fastllve} + VSR~\cite{xu2024enhancing} methods; (c) SOTA one-stage LVSR method~\cite{lu2023learning}; and (e) our event-based approach. It can be observed that only our method produces well-lit, high-quality results with clearly recognizable text.}
        \label{fig:fig1}
        \vspace{-1ex}
    \end{figure}

    To achieve VSR from low-light videos,~\ie, low-light VSR (LVSR), a straightforward approach is to first apply low-light video enhancement (LVE)~\cite{li2023fastllve}, followed by VSR methods, which we refer to as the \textit{cascade} strategy. However, this approach has a major drawback in that the pixel errors introduced during the LVE stage are propagated and amplified in the VSR step, thus degrading the overall performance. An alternative strategy is to perform VSR first and then apply LVE. However, the quality deteriorates because the super-resolved frames suffer from weakened textures, amplified noise, and low contrast. To address these issues, ~\citet{xu2023deep} proposed the first one-stage LVSR model that directly learns a mapping from low-light LR inputs to well-lit HR outputs. However, as shown in Fig.~\ref{fig:fig1}, these methods still suffer from severe artifacts, structural distortions, and inaccurate illumination.

    LVSR is a very challenging problem. It is difficult to rely solely on low-light LR frames to restore high-quality HR videos due to the inherent lack of sufficient contrast to distinguish fine textures, as well as the lack of high-frequency details in LR frames. In addition, sudden lighting changes at night, such as flashes from streetlights or car headlights, further exacerbate the problem. Recently, event signals captured by event cameras have been used for low-light enhancement~\cite{liang2023coherent}, super-resolution~\cite{kai2023video}, and high dynamic range imaging~\cite{han2023hybrid}. Compared with standard cameras, event cameras offer a very high dynamic range (120 dB), high temporal resolution (about 1~$\mu s$), and rich ``moving edge" information~\cite{gallego2020event}. These characteristics enable event signals to provide complementary cues, such as sharp edges and motion details, even at night, for LVSR. Motivated by these advantages, we propose including event signals as auxiliary information to improve LVSR performance.

\begin{figure}[t!]
        \centering
        \includegraphics[width=\columnwidth]{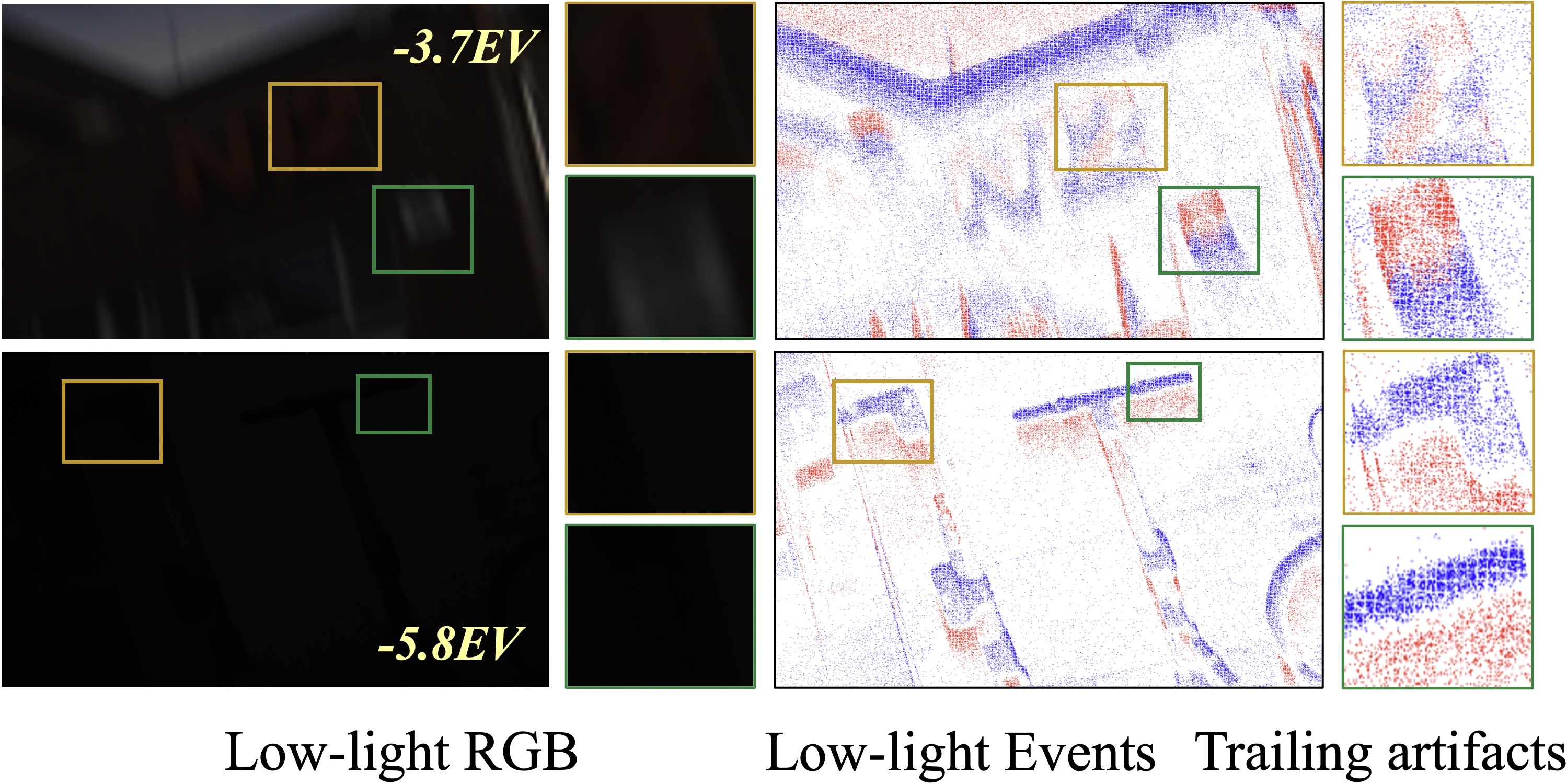}
        \caption{In low light, both RGB and event signals degrade: the RGB frame suffers from severe illumination and detail loss, and the event data contains noise and trailing artifacts.}
        \label{fig:fig2}
        \vspace{-1ex}
    \end{figure}

    However, while event signals offer valuable information, effectively integrating them into LVSR remains challenging. As shown in Fig.~\ref{fig:fig2}, not only are RGB frames heavily degraded under low-light conditions, but event data also suffers from noise, temporal trailing effects, and spatially non-stationary distributions~\cite{liu2025ner}. Directly fusing such degraded event signals with low-quality RGB frames inevitably introduces noise and artifacts into the reconstructed results. Therefore, how to effectively extract and fuse meaningful information from both degraded signals is of paramount importance for event-based LVSR.

\begin{figure}[t!]
        \centering
        \includegraphics[width=\columnwidth]{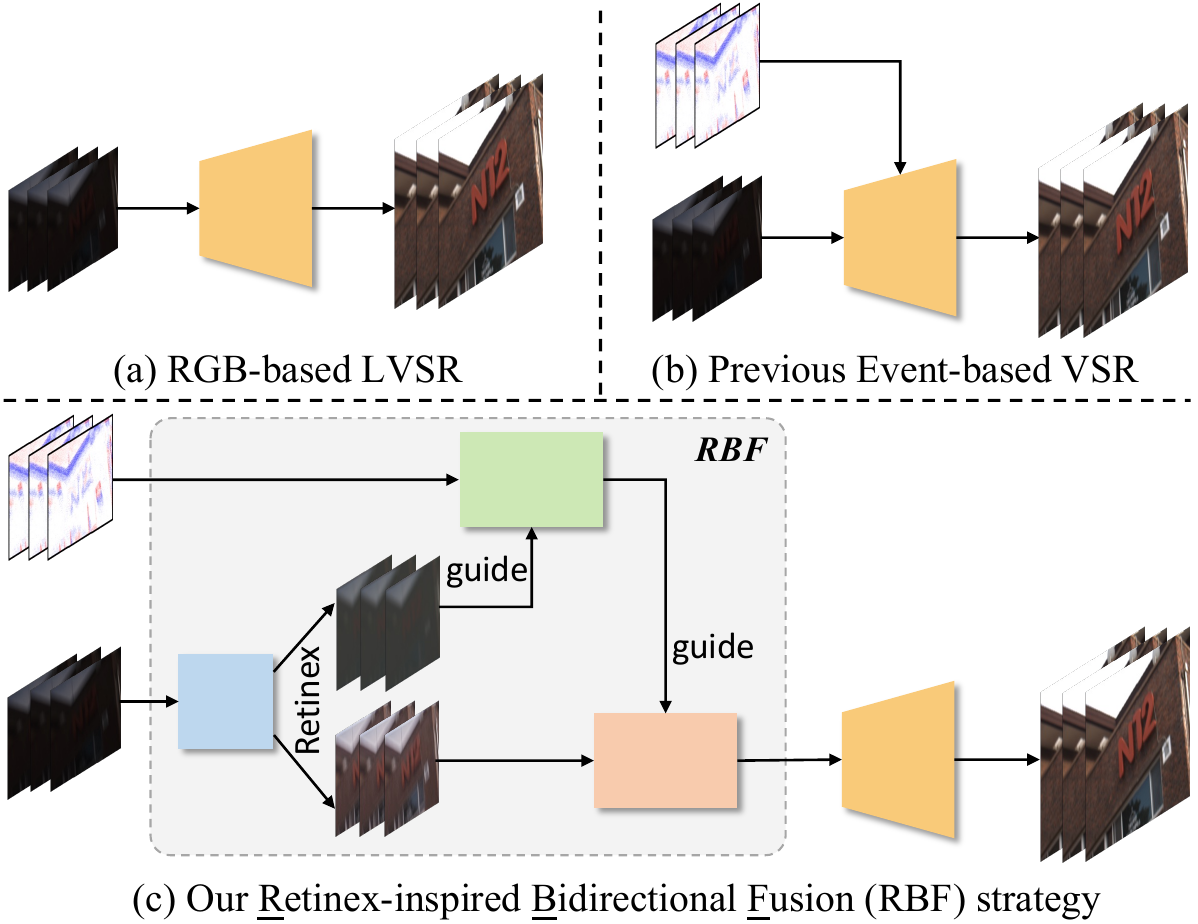}
        \caption{Comparison of LVSR strategies. (a) RGB-based  method~\cite{xu2023deep} directly super-resolves low-light frames. (b) Previous event-based methods~\cite{lu2023learning,kai2024evtexture} directly fuse two degraded modalities. (c) Our RBF strategy first uses illumination to guide event refinement and then leverages the refined events to enhance reflectance, enabling effective information integration.}
        \label{fig:fig3}
        \vspace{-1ex}
    \end{figure}

    To achieve this, we first argue that the degradation in both modalities mainly arises from insufficient lighting, and that relying solely on event data is inadequate to address these issues without additional low-light priors. To address this, we draw inspiration from Retinex decomposition~\cite{wei2018deep}, which separates a low-light image into illumination and reflectance. Illumination provides smooth, low-noise global lighting cues, while reflectance preserves intrinsic scene content but lacks fine details in LR settings. Based on this insight, we propose a Retinex-inspired Bidirectional Fusion (RBF) strategy: illumination guides the refinement of noisy events, and enhanced events are then used to recover reflectance details, as illustrated in Fig.~\ref{fig:fig3}(c). This bidirectional process enables effective mutual guidance between RGB and event modalities.

    To this end, we present RetinexEVSR, an innovative LVSR network that integrates high-contrast event signals with Retinex-inspired priors to enhance video quality under low-light conditions. In our RetinexEVSR, the input frames are first decomposed into illumination and reflectance components. Guided by the proposed RBF strategy, we introduce an Illumination-guided Event Enhancement (IEE) module, which progressively refines event features through multi-scale fusion with illumination, enabling hierarchical guidance from coarse to fine levels. The refined events are then passed to the Event-guided Reflectance Enhancement (ERE) module to recover reflectance details. This module adopts a dynamic attention mechanism to inject high-frequency information from events into the reflectance stream via multi-scale fusion. Finally, the illumination, enhanced reflectance, and refined event features are jointly used to guide the upsampling process, reducing information loss and improving reconstruction quality. Experimental results on three datasets demonstrate the effectiveness of our proposed RetinexEVSR, which remains robust even under extreme darkness and severe motion blur. To summarize, our main contributions are:
    \begin{itemize}
        \item We present RetinexEVSR, the first event-driven scheme for LVSR. Our RetinexEVSR leverages event signals and Retinex-inspired priors to restore severely degraded RGB inputs under low-light conditions.
        \item We introduce a novel RBF strategy to enable effective cross-modal fusion between RGB and event signals, addressing the challenge of combining degraded inputs.
        \item We propose the IEE and ERE modules to progressively enhance event and reflectance features, enabling coarse-to-fine guidance and detailed texture restoration.
        \item RetinexEVSR achieves state-of-the-art performance on three datasets, including synthetic and real-world data.
    \end{itemize}

    \begin{figure*}[t!]
        \centering
        \includegraphics[width=\textwidth]{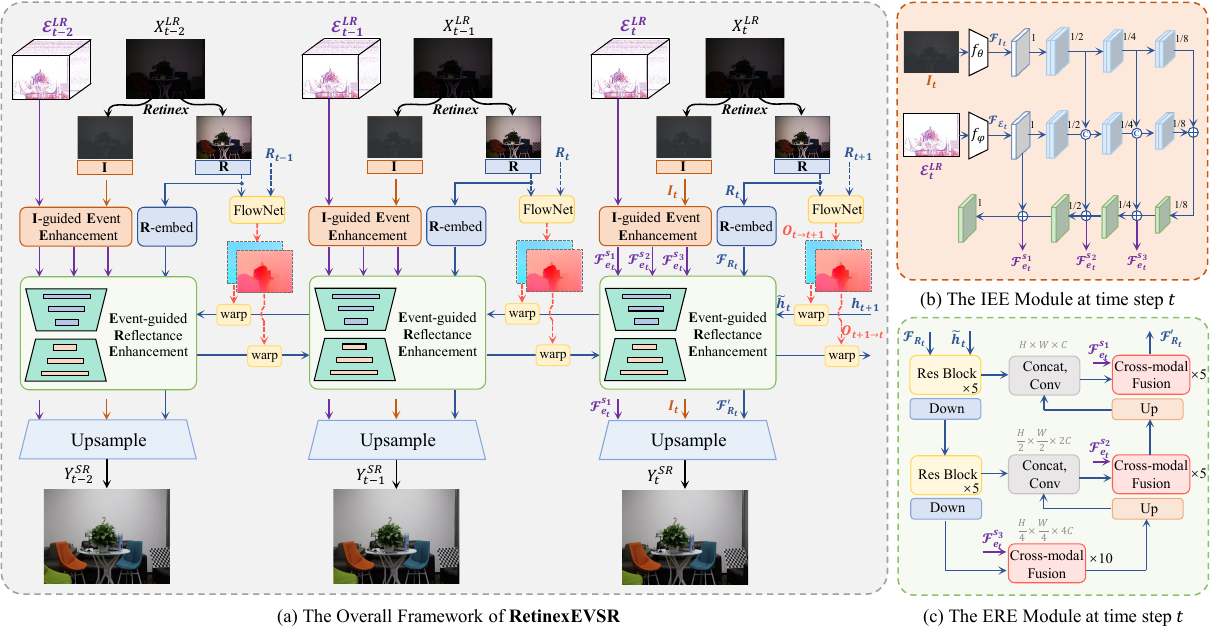}
        \caption{Network architecture of RetinexEVSR. (a) The model takes low-light LR frames and corresponding events as input, and outputs HR frames with well-lit details. Each frame is decomposed into illumination and reflectance, and optical flow is estimated from reflectance for temporal alignment. (b) At each time step, the IEE module uses illumination to guide event enhancement. (c) The refined event features are then used in the ERE module to enhance reflectance features.}
        \label{fig:fig4}
        \vspace{-1ex}
    \end{figure*}

    \section{Related Work}
    \label{sec:related_work}

    \paragraph{Video Super-Resolution.} As a fundamental computer vision task, VSR technology has made remarkable progress in recent years~\cite{li2025ustc,wei2025evenhancer,Xie_2025_ICCV}. The essential challenge in VSR is to predict the missing details of the current HR frame from other unaligned frames. To achieve this, many advanced alignment~\cite{tang2024semantic} and propagation~\cite{du2025patchvsr} methods have been devised. However, these methods often perform poorly under low-light conditions due to issues such as amplified noise and weakened textures. To address this, some works~\cite{xu2023joint,gao2024dual} have proposed joint learning of low-light enhancement and super-resolution.~\citet{xu2023deep} introduced the first one-stage LVSR framework that directly learns a mapping from low-light LR videos to normal-light HR videos. However, the method still struggles with large displacements and motion blur, resulting in severe temporal inconsistency.
    
    \paragraph{Low-Light Video Enhancement.} To achieve LVE, a common strategy is to apply low-light image enhancement (LIE) methods to each frame independently. In recent years, a large number of CNN-based~\cite{wu2025codebook,wu2025lightqanet,ju2025illumination} and Transformer-based~\cite{wang2023ultra,cai2023retinexformer} LIE methods have emerged. Among them, the Retinex model~\cite{wei2018deep} is a popular tool for LIE, where an observed image $X$ can be expressed as $X=R \odot I$. Here, $R$ and $I$ represent reflectance and illumination maps, respectively, and $\odot$ denotes element-wise multiplication.~\citet{cai2023retinexformer} introduced Retinexformer, the first Transformer-based method that uses illumination derived from Retinex theory to guide the modeling of long-range dependencies in the self-attention mechanism. 

    However, frame-by-frame LIE often causes temporal flickering and jitter effects due to dynamic illumination changes in low-light videos. To address these, many one-stage LVE methods~\cite{zhu2024temporally,zhu2024unrolled} have been proposed.~\citet{li2023fastllve} proposed an efficient pipeline named FastLLVE, which leverages the look-up table technique to effectively maintain inter-frame brightness consistency. However, they still face limitations in using temporal redundancy in low-light videos due to difficulties in extracting distinct features for motion estimation.

    \newcounter{typecounter}
\setcounter{typecounter}{1}

\begin{table*}[t!]
    \centering
    \resizebox{\textwidth}{!}{  
    \begin{tabular}{cl|ccc|ccc|ccc|ccc}
        \toprule[0.15em]
        \multirow{2}[1]{*}{Type} & \multirow{2}[1]{*}{Method} & \multicolumn{3}{c|}{SDSD-in} & \multicolumn{3}{c|}{SDSD-out} & \multicolumn{3}{c|}{SDE-in} & \multicolumn{3}{c}{SDE-out} \\
        & & PSNR$\uparrow$ & SSIM$\uparrow$ & LPIPS$\downarrow$ & PSNR$\uparrow$ & SSIM$\uparrow$ & LPIPS$\downarrow$ & PSNR$\uparrow$ & SSIM$\uparrow$ & LPIPS$\downarrow$ & PSNR$\uparrow$ & SSIM$\uparrow$ & LPIPS$\downarrow$ \\
        \midrule
    
        \multirow{4}[1]{*}{\Roman{typecounter}} & Retinexformer $+$ MIA-VSR & 27.11 & 0.8341 & 0.3962 & 19.81 & 0.6634 & 0.5051 & 16.68 & 0.4053 & 05533 & 16.20 & 0.3981 & 0.5507 \\
        & FastLLVE $+$ MIA-VSR & 16.32 & 0.7224 & 0.5120 & 20.22 & 0.6500 & 0.5598 & 13.91 & 0.3300 & 0.6367 & 13.79 & 0.3030 & 0.6452 \\
        & Retinexformer $+$ IART & 27.09 & 0.8331 & 0.3938 & 19.84 & 0.6638 & 0.5051 & 16.68 & 0.4053 & 0.5533 & 16.17 & 0.3967 & 0.5473 \\
        & FastLLVE $+$ IART & 16.33 & 0.7217 & 0.5095 & 20.25 & 0.6508 & 0.5578 & 13.91 & 0.3300 & 0.6373 & 13.77 & 0.3011 & 0.6407 \\
        \midrule
    
        \addtocounter{typecounter}{1}
        \multirow{4}[1]{*}{\Roman{typecounter}} & MIA-VSR $+$ Retinexformer & 25.10 & 0.8457 & 0.3644 & 23.80 & 0.7583 & 0.4316 & 16.82 & 0.4607 & 0.4716 & 15.94 & 0.4011 & 0.4942 \\
        & MIA-VSR $+$ FastLLVE & 23.94 & 0.8261 & 0.4033 & 13.71 & 0.5364 & 0.4303 & 16.39 & 0.4875 & 0.5028 & 15.80 & 0.4104 & 0.5200 \\
        & IART $+$ Retinexformer & 25.30 & 0.8500 & 0.3541 & 24.07 & 0.7566 & 0.4192 & 17.73 & 0.4339 & 0.4952 & 16.00 & 0.4088 & 0.4799 \\
        & IART $+$ FastLLVE & 24.03 & 0.8304 & 0.3944 & 13.61 & 0.5345 & 0.4261 & 16.38 & 0.4869 & 0.5017 & 15.75 & 0.4108 & 0.5105 \\
        \midrule
    
    
        \addtocounter{typecounter}{1}
        \multirow{4}[1]{*}{\Roman{typecounter}} & EvLight $+$ EGVSR & 26.57 & 0.8220 & 0.3944 & 20.59 & 0.6752 & 0.4214 & 19.61 & 0.5939 & 0.5120 & 19.21 & 0.5296 & 0.5198 \\
        & EvLowLight $+$ EGVSR & 18.61 & 0.6783 & 0.4638 & 14.75 & 0.5147 & 0.5465 & 19.81 & 0.5626 & 0.5923 & 17.46 & 0.4389 & 0.6393 \\
        & EvLight $+$ EvTexture & 26.15 & 0.8127 & 0.3823 & 19.65 & 0.6504 & 0.4214 & 19.54 & 0.5700 & 0.5070 & 19.58 & 0.5230 & .5227 \\
        & EvLowLight $+$ EvTexture & 18.46 & 0.6452 & 0.4634 & 14.52 & 0.4615 & 0.5545 & 19.32 & 0.5118 & 0.5813 & 17.52 & 0.4187 & 0.6312 \\
        \midrule
    
    
        \addtocounter{typecounter}{1}
        \multirow{4}[1]{*}{\Roman{typecounter}} & EGVSR $+$ CoLIE & 13.53 & 0.6844 & 0.3888 & 23.39 & 0.7351 & 0.4064 & 15.27 & 0.3096 & 0.4963 & 14.50 & 0.2404 & 0.5151 \\
        & EGVSR $+$ Zero-IG & 16.86 & 0.6904 & 0.4533 & 9.77 & 0.3957 & 0.4934 & 19.06 & 0.5056 & 0.5491 & 18.03 & 0.4607 & 0.5559 \\
        & EvTexture $+$ CoLIE & 9.46 & 0.2671 & 0.5767 & 23.12 & 0.7457 &  \underline{0.4014} & 15.27 & 0.3096 & 0.4963 & 14.43 & 0.2404 & 0.4941 \\
        & EvTexture $+$ Zero-IG & 10.85 & 0.3747 & 0.5755 & 9.78 & 0.4006 & 0.4680 & 19.06 & 0.5056 & 0.5491 & 18.16 & 0.4723 & 0.5337 \\
        \midrule
    
        \addtocounter{typecounter}{1}
        \multirow{5}[1]{*}{\Roman{typecounter}} & BasicVSR++  & 25.90 & 0.8496 & 0.3481 & 22.87 & 0.7115 & 0.4200 & 19.91 & 0.6128 & 0.5123 & 19.44 & 0.5768 & 0.5307 \\
        & DP3DF & 27.23  & 0.8445  & 0.3413  & 22.25  & 0.7299  & 0.4161  & 16.99  & 0.4007  & 0.5122  & 14.89  & 0.3523  & 0.5114  \\
        & MIA-VSR  & 15.71 & 0.6619 & 0.4863 & 19.57 & 0.6777 & 0.4545 & 19.06 & 0.5284 & 0.5492 & 16.82 & 0.4834 & 0.5703 \\
        & IART  & 23.74 & 0.8331 & 0.3699 & 24.15 & 0.7400 & 0.4260 & 19.33 & 0.5671 & 0.5008 & 19.82 & 0.6109 & 0.4886 \\
        & FMA-Net  &  \underline{27.53} & 0.8680 & 0.3300 & 23.93 & 0.7473 & 0.4084 &  \underline{20.31} &  \underline{0.6334} &   \textbf{0.4578} & 19.86 &  \underline{0.6156} &  \underline{0.4709} \\
        \midrule
    
        \addtocounter{typecounter}{1}
        \multirow{3}[1]{*}{\Roman{typecounter}} & EGVSR  & 27.24 & 0.8559 & 0.3698 & 23.71 & 0.7483 & 0.4167 & 19.78 & 0.5780 & 0.5734 &  \underline{19.92} & 0.5716 & 0.5256 \\
        & EvTexture  & 27.33 &  \underline{0.8776} &  \underline{0.3286} &  \underline{24.20} &  \underline{0.7587} & 0.4166 & 20.29 & 0.6301 & 0.4869 & 19.75 & 0.6046 & 0.4977 \\
        & \textbf{RetinexEVSR (Ours)} &   \textbf{30.28} &   \textbf{0.8932} &   \textbf{0.3149} &   \textbf{25.15} &   \textbf{0.7737} &   \textbf{0.3933} &   \textbf{21.24} &   \textbf{0.6525} &  \underline{0.4627} &   \textbf{20.68} &   \textbf{0.6541} &   \textbf{0.4382} \\
        \bottomrule[0.15em]
    \end{tabular}
    }
    \caption{Quantitative comparison on SDSD and SDE datasets for $4\times$ LVSR. All methods are retrained on the same dataset. All results are calculated on the RGB channel. \textbf{Bold} and \underline{underlined} numbers indicate the best and second-best performance.}
    \label{tab:table1}
    \end{table*}

    \begin{table*}[t!]
\centering
\resizebox{\textwidth}{!}{  
\begin{tabular}{l|cc|cc|ccccccc}
    \toprule[0.15em]
     \multirow{3}[2]{*}{Method} & \multicolumn{2}{c|}{Enhancement $+$ VSR} & \multicolumn{2}{c|}{VSR $+$ Enhancement} & \multicolumn{6}{c}{Joint Enhancement and VSR} \\
     \cmidrule{2-3} \cmidrule{4-5} \cmidrule{6-12}
     & \multirow{2}[1]{*}{\makecell{Retinexformer\\$+$ IART}} & \multirow{2}[1]{*}{\makecell{EvLight\\$+$ EvTexture}} & \multirow{2}[1]{*}{\makecell{MIA-VSR\\$+$ FastLLVE}} & \multirow{2}[1]{*}{\makecell{EGVSR\\$+$ CoLIE}}  & \multirow{2}[1]{*}{DP3DF} & \multirow{2}[1]{*}{MIA-VSR} & \multirow{2}[1]{*}{IART} & \multirow{2}[1]{*}{FMA-Net} & \multirow{2}[1]{*}{EGVSR} & \multirow{2}[1]{*}{EvTexture} & \multirow{2}[1]{*}{\textbf{Ours}} \\[1.3em]
    \midrule
     PSNR$\uparrow$ & 15.43 & 17.22 & 19.30 & 18.76 & 27.02 & 24.48 & 26.36 & 27.61 & 26.90 &   \underline{28.07} &    \textbf{28.92} \\
     SSIM$\uparrow$ & 0.5861 & 0.6506 & 0.7123 & 0.7462 & 0.8406 & 0.8194 & 0.8402 &   \underline{0.8611} & 0.8473  & 0.8604 &    \textbf{0.8707} \\
     LPIPS$\downarrow$ & 0.5749 & 0.5760 & 0.5834 & 0.5073 &   \underline{0.4625} & 0.5383 & 0.4731 & 0.4633 & 0.5060 & 0.4837 &    \textbf{0.4612}\\
     tOF$\downarrow$ & 6.55 & 8.87 & 6.76 & 6.07 & 5.65 & 5.86 & 5.93 & 4.70 & 5.27 &   \underline{4.69} &    \textbf{4.60} \\
     TCC$\uparrow$\scalebox{0.5}{$ \times  10$} & 1.50 & 1.26 & 0.80 & 1.88 & 2.88 & 2.40 & 2.85 &   \underline{3.17} & 2.96 & 3.14 &    \textbf{3.31} \\
     \midrule
     Params (M) & 1.61+13.41 & 22.73+8.90  & 16.60+11.11 & 2.58+0.13 & 28.86 & 16.60 & 13.41 & 9.62 & \textbf{2.58} & 8.90 & \underline{8.07} \\
     FLOPs (G) & 21.3+2778.4 & 241.5+1141.1 & 1755.5+87.0  & 226.9+8.7 & 775.3 & 1755.5 & 2778.4 & 1941.3 & \underline{226.9} & 1141.1 & \textbf{159.1} \\
     Runtime (ms) & 17.3+1666.8 & 50.8+126.9 & 1159.1+28.6 & 181.7+7.5 & \underline{52.7} & 1159.1 & 1666.8 & 596.3 & 181.7 & 126.9 & \textbf{44.5} \\  
    \bottomrule[0.15em]
\end{tabular}
}
\caption{Quantitative comparison on RELED for $4\times$ LVSR. FLOPs and runtime are computed on one $256\times320$ LR frame.}
\label{tab:table2}
\end{table*}

    \paragraph{Event-based Vision.} Event cameras are bio-inspired sensors that offer several advantages over standard RGB cameras, including ultra-high temporal resolution (about $1\mu$s)~\cite{xiao2024estme}, high dynamic range (120 dB), and low power (5 mW). They have been widely used for tasks like frame interpolation~\cite{Liu_2025_CVPR,sun2025exploring,liu2025nemf}, deblurring~\cite{yang2024learning,yang2025asymmetric}, and low-light enhancement~\cite{zhang2024sim,kim2024towards}.
    
    More closely related to our work, recent studies~\cite{xiao2024event,xiao2024asymmetric,yan2025evstvsr,kai2025event,xiao2025event} have introduced event signals for VSR. For instance,~\citet{jing2021turning} proposed the first event-based VSR method, named E-VSR, which uses events for frame interpolation followed by VSR, enhancing overall performance.~\citet{kai2024evtexture} introduced EvTexture, utilizing high-frequency information from events to improve texture restoration. While these methods perform well under normal-light conditions, they struggle in low-light scenarios. The challenge of training with event data for VSR under low-light conditions remains largely unexplored.

    \section{Method}
    
    \paragraph{RetinexEVSR Framework.}
    We propose a novel neural network, named RetinexEVSR, to address the challenge of VSR under low-light conditions by leveraging high-contrast event signals and Retinex-inspired priors. The architecture of RetinexEVSR is illustrated in Fig.~\ref{fig:fig4}(a). The input consists of a LR image sequence $\{X^{LR}_t\}_{t=1}^{T}$ with $T$ frames and the corresponding event data $\{\mathcal{E}^{LR}_t\}_{t=1}^{T}$. The network outputs a super-resolved, well-lit image sequence $\{Y^{SR}_t\}_{t=1}^{T}$.

    At a given time step $t$, the input frame $X^{LR}_t$ is first decomposed into illumination $I_t$ and reflectance $R_t$ via a Retinex-based LIE model, such as SCI~\cite{ma2022toward,ma2025learning}. 
    The illumination $I_t$ is used to guide event feature extraction within the IEE module, producing multi-scale event features. In our implementation, we use three scales: $\{\mathcal{F}^{s_1}_{e_t}, \mathcal{F}^{s_2}_{e_t}, \mathcal{F}^{s_3}_{e_t}\}$, where $s_1$ corresponds to the largest spatial scale. The reflectance $R_t$ is fed into the R-embed layer, which consists of five Residual Blocks adopted from~\cite{wang2018esrgan}, to extract the feature representation $\mathcal{F}_{R_t}$. This feature is then enhanced by events in the ERE module, yielding the enhanced reflectance feature $\mathcal{F}'_{R_t}$. Finally, features from events, illumination, and reflectance are fused to guide upsampling, producing the final output $Y^{SR}_t$. 

    \begin{figure*}[t!]
        \centering
        \includegraphics[width=\textwidth]{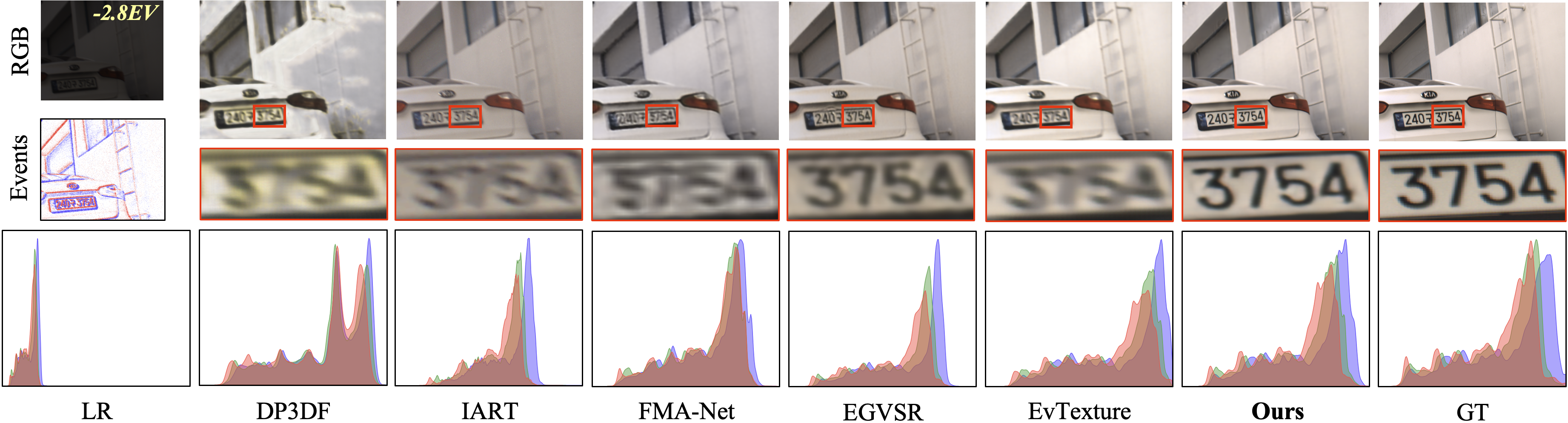}
        \caption{Qualitative comparison on RELED for 4$\times$ LVSR. The bottom row is the statistical distribution of the RGB channels. Our method recovers clearer license plate numbers and more faithful colors that better match the ground truth.}
        \label{fig:fig5}
    \end{figure*}

    \begin{figure*}[t!]
        \centering
        \includegraphics[width=\textwidth]{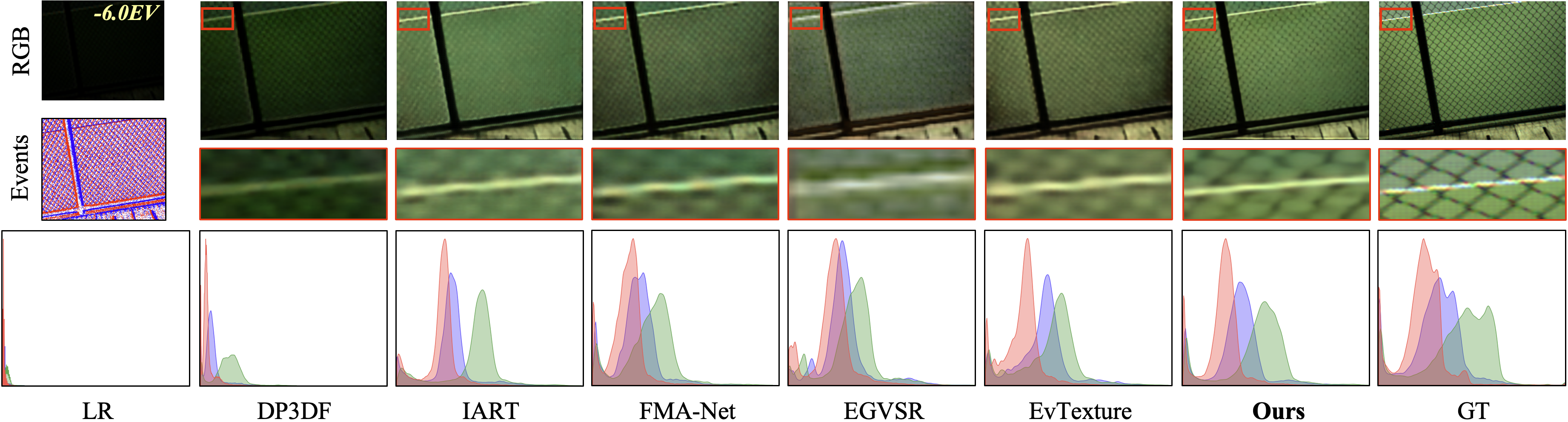}
        \caption{Qualitative comparison on SDE for 4$\times$ LVSR.  Our method effectively restores well-lit images with fine details.}
        \label{fig:fig6}
        \vspace{-1ex}
    \end{figure*}

    From a temporal perspective, RetinexEVSR employs a bidirectional recurrent framework~\cite{chan2021basicvsr}, where inter-frame optical flow serves as a \textit{bridge} for temporal alignment and feature propagation. Unlike prior methods~\cite{xu2023deep}, we compute flow from reflectance maps instead of raw inputs, as they offer higher contrast and enable more accurate alignment under low-light conditions. For example, between timestamps $t$ and $t+1$, flows $O_{t+1 \to t}$ and $O_{t \to t+1}$ are computed between $R_t$ and $R_{t+1}$. In backward propagation, the feature $h_{t+1}$ is warped to time $t$ using $O_{t \to t+1}$ via a backward warping operation, producing the aligned feature $\tilde{h}_t$, which is then fed into the ERE module for reflectance enhancement.

    \paragraph{Illumination-guided Event Enhancement.}
    Under low-light conditions, both event signals and RGB frames suffer significant degradation. Directly fusing them for LVSR often leads to artifacts due to the compounded noise and distortions from both modalities. To address this, we propose the IEE module, which leverages the illumination map as a global lighting prior to \textit{light up} the event feature extraction and suppress low-light noise. As shown in Fig.~\ref{fig:fig4}(b), at time step $t$, given the illumination map $I_t$ and LR event $\mathcal{E}^{LR}_t$, the IEE module first extracts shallow features using two symmetric branches: $f_\theta$ for illumination and $f_\varphi$ for events. Both branches adopt lightweight residual blocks:
    \begin{equation}
        \mathcal{F}_{I_t} = f_\theta(I_t), \quad \mathcal{F}_{\mathcal{E}_t} = f_\varphi(\mathcal{E}^{LR}_t),
    \end{equation}
    where $\mathcal{F}_{I_t}$ and $\mathcal{F}_{\mathcal{E}_t}$ denote the initial features from illumination and events. However, $\mathcal{F}_{\mathcal{E}_t}$ still suffers from trailing artifacts and noise. To refine event features, we adopt a multi-scale fusion strategy inspired by~\cite {wang2023lrru}. Convolutions with varying kernel sizes are used to extract features at four spatial scales: full, half, quarter, and one-eighth resolution (\ie, $1$, $1/2$, $1/4$, and $1/8$), enabling the network to perceive illumination-aware cues across multiple receptive fields. At each scale, illumination features guide the fusion process via channel-wise concatenation and convolution, allowing the network to recalibrate event representations based on lighting priors. The fused features are then progressively upsampled from coarse to fine in a top-down refinement pathway. At each stage, they are combined with finer-scale event features to recover spatial details while maintaining illumination consistency. We retain the top three scales after fusion as the final enhanced event features: $\{\mathcal{F}^{s_1}_{e_t}, \mathcal{F}^{s_2}_{e_t}, \mathcal{F}^{s_3}_{e_t}\}$, where $s_1$ corresponds to the largest spatial scale. This hierarchical strategy effectively enhances event representations under low-light conditions, providing reliable guidance for subsequent reconstruction.

    \paragraph{Event-guided Reflectance Enhancement.}
    In Retinex-based LIE, reflectance is commonly used as the target since it carries well-lit content and structural information. However, in the LVSR setting, it often lacks high-frequency details. To compensate for this, we propose the ERE module, which utilizes refined event features—enhanced by the IEE module—to supplement reflectance features with high-frequency cues. As illustrated in Fig.~\ref{fig:fig4}(c), the ERE module adopts an `encoder–bottleneck–decoder' architecture. To incorporate temporal information, we introduce the temporally propagated feature $\widetilde{h}_t$ into the input. Additionally, event and reflectance features are dynamically fused in both the bottleneck and decoder stages through an attention-based cross-modal fusion~\cite{li2024contourlet} block.  This design enables the network to selectively inject informative structures from events into the reflectance stream while suppressing noise specific to either modality. After processing through the ERE module, the original reflectance feature $\mathcal{F}_{R_t}$ is enriched with detailed textures and contrast information from the event features. The output, denoted as $\mathcal{F}^{'}_{R_t}$, also serves as the updated temporal feature $h_t$ for the next frame, enabling continuous refinement. This enhancement not only improves the perceptual quality of the reconstructed frames but also provides stronger guidance for the final restoration. Further details about the fusion block are provided in the appendix.
    
    \begin{table}[t!]
	\centering
	\resizebox{\columnwidth}{!}{  
		\begin{tabular}{l|l|cccc}
			\toprule[0.15em]
                Datasets & Methods & NIQE$\downarrow$ & PI$\downarrow$ &  CLIP-IQA$\uparrow$ &
                Q-Align$\uparrow$\\
                \midrule
			\multirow{5}[1]{*}{\makecell{SDE-in}} & DP3DF &    \textbf{6.8206} & 7.5631  & 0.1540  & 1.3184  \\
            & IART & 10.6221 & 9.0256 & 0.1889 & 1.3207 \\
            & EGVSR & 9.0954  & 8.1349  &   \textbf{0.3063}  &  \underline{1.4150}  \\
            & EvTexture & 8.5623 &  \underline{7.4788}  & 0.1993  & 1.2285  \\
            & \multicolumn{1}{>{\columncolor[gray]{0.8}}l|}{\textbf{Ours}} & 
\multicolumn{1}{>{\columncolor[gray]{0.8}}c}{ \underline{7.0684}} & 
\multicolumn{1}{>{\columncolor[gray]{0.8}}c}{\textbf{  7.2035}} & 
\multicolumn{1}{>{\columncolor[gray]{0.8}}c}{ \underline{0.2588}} & 
\multicolumn{1}{>{\columncolor[gray]{0.8}}c}{  \textbf{1.6426}} \\
            \midrule
            \midrule
            \multirow{5}[1]{*}{\makecell{SDE-out}} & DP3DF &  7.5242 & 7.1929  & 0.1510  & 1.2910 \\
            & IART &  \underline{7.1097}  &  \underline{7.1529}  & 0.1342  & 1.5479  \\
            & EGVSR & 9.6327 & 8.5254  &  \underline{0.2537}  & 1.5928  \\
            & EvTexture & 8.0480  & 8.4553  & 0.2377  &  \underline{1.6209}  \\
            & \multicolumn{1}{>{\columncolor[gray]{0.8}}l|}{\textbf{Ours}} & 
\multicolumn{1}{>{\columncolor[gray]{0.8}}c}{  \textbf{6.7292}} & 
\multicolumn{1}{>{\columncolor[gray]{0.8}}c}{  \textbf{7.0141}} & 
\multicolumn{1}{>{\columncolor[gray]{0.8}}c}{  \textbf{0.2618}} & 
\multicolumn{1}{>{\columncolor[gray]{0.8}}c}{  \textbf{1.7432}} \\
            \bottomrule[0.15em]
		\end{tabular}
	}
        \caption{Generalization to real-world SR on the SDE dataset. }
	\label{tab:table3}
\end{table}

    \begin{table}[t!]
	\centering
	\resizebox{\columnwidth}{!}{  
		\begin{tabular}{cclcccc}
			\toprule[0.15em]
			\multicolumn{3}{c}{\multirow{2}[1]{*}{Method}} & \multicolumn{3}{c}{SDSD-in} & \multirow{2}[1]{*}{\makecell{\#Params \\ (M)}} \\ 
			& & & PSNR$\uparrow$ & SSIM$\uparrow$ & LPIPS$\downarrow$  & \\
			\midrule
			\multirow{3}[1]{*}{\makecell{Break\\-down}} & &  (a) w/o IEE & 28.27 & 0.8642  & 0.3274 & 7.26 \\
                & &  (b) w/o ERE & 27.31 & 0.8422 & 0.3304 & 6.38 \\
                &  $*$ &  (c) Full Model &   30.28 &   0.8932 &   0.3149 & 8.07 \\
			\midrule
			\multirow{3}[1]{*}{IEE} & & (d) $scale=1$ & 28.64 & 0.8772 & 0.3313 & 7.28 \\
			& & (e) $scale=2$ & 28.83 & 0.8801 & 0.3239 & 7.71 \\
                &  $*$ & (f) $scale=3$ &   30.28 &   0.8932 &   0.3149 & 8.07 \\
			\midrule
			\multirow{2}[0]{*}{ERE} & & (g) single-scale & 28.04 & 0.8553 & 0.3325 & 8.05 \\
			& &  (h) w/o fusion & 29.78  & 0.8911 & 0.3172 & 8.06 \\
			\midrule
			\multirow{2}[0]{*}{\makecell{Retinex\\Model}} & &  (i) URetinex & 29.62 & 0.8873 & 0.3294 & 8.09 \\
			&  $*$  &  (j) SCI &   30.28 &   0.8932 &   0.3149 & 8.07 \\
			\midrule
			\multirow{2}[0]{*}{\makecell{Optical\\Flow}} & &  (k) from $\{X_t\}$ & 29.85 & 0.8908 & 0.3196 & 8.07 \\
			&  $*$  &  (l) from $\{R_t\}$ &   30.28 &   0.8932 &   0.3149 & 8.07 \\
			\bottomrule[0.15em]
		\end{tabular}
	}
        \caption{Ablation study of model components on SDSD-indoor. $*$ indicates the setting used in our final model.}
	\label{tab:table4}
  \vspace{-1ex}
\end{table}

    \paragraph{Loss Function.}
    We follow the previous study~\cite{kai2024evtexture} and adopt the Charbonnier loss~\cite{lai2017deep} as the training loss function, which is defined as:
    \begin{equation}
    \mathcal{L} = \frac{1}{T} \sum_{t=1}^{T} \sqrt{\left\|Y_{t}^{GT} - Y_{t}^{SR}\right\|^2 + \varepsilon^2},
    \end{equation}
    where $\varepsilon=1 \times 10^{-12}$ is set for numerical stability.

    \section{Experiments}
    \noindent\textbf{Datasets.} We first follow the previous LVSR method DP3DF~\cite{xu2023deep} and use the SDSD dataset~\cite{wang2021seeing}, which provides paired low-light and normal-light videos. Since SDSD does not include event signals, we simulate events using the vid2e event simulator~\cite{gehrig2020video} with a noise model based on ESIM~\cite{rebecq2018esim}.
    We further train and evaluate our method on two real-world event datasets: SDE~\cite{liang2024towards} and RELED~\cite{kim2024towards}. SDE contains over 30K image-event pairs captured under varying lighting conditions in indoor and outdoor scenes. RELED introduces severe motion blur caused by long exposures in low-light, making it more challenging.
    Event data is converted into voxel grids~\cite{zhu2021eventgan} with 5 temporal bins and downsampled using the same bicubic interpolation as the frames.

    \begin{figure}[t!]
    	\centering
    	\includegraphics[width=\columnwidth]{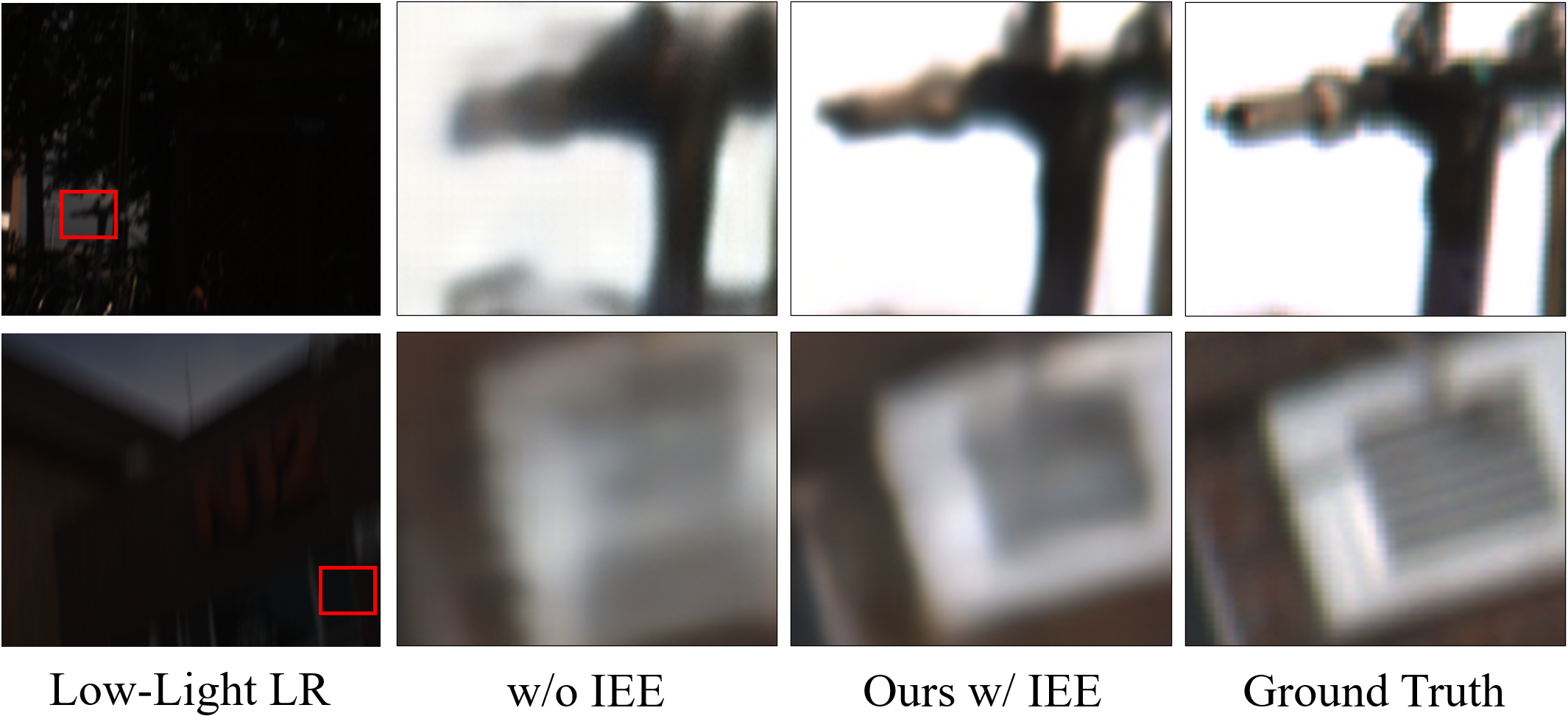}
        \caption{Ablation study of IEE. The full model produces sharper structures and finer details.}
    	\label{fig:fig7}
        \vspace{-0.5ex}
    \end{figure}
    
    \begin{figure}[t!]
        \centering
        \includegraphics[width=\columnwidth]{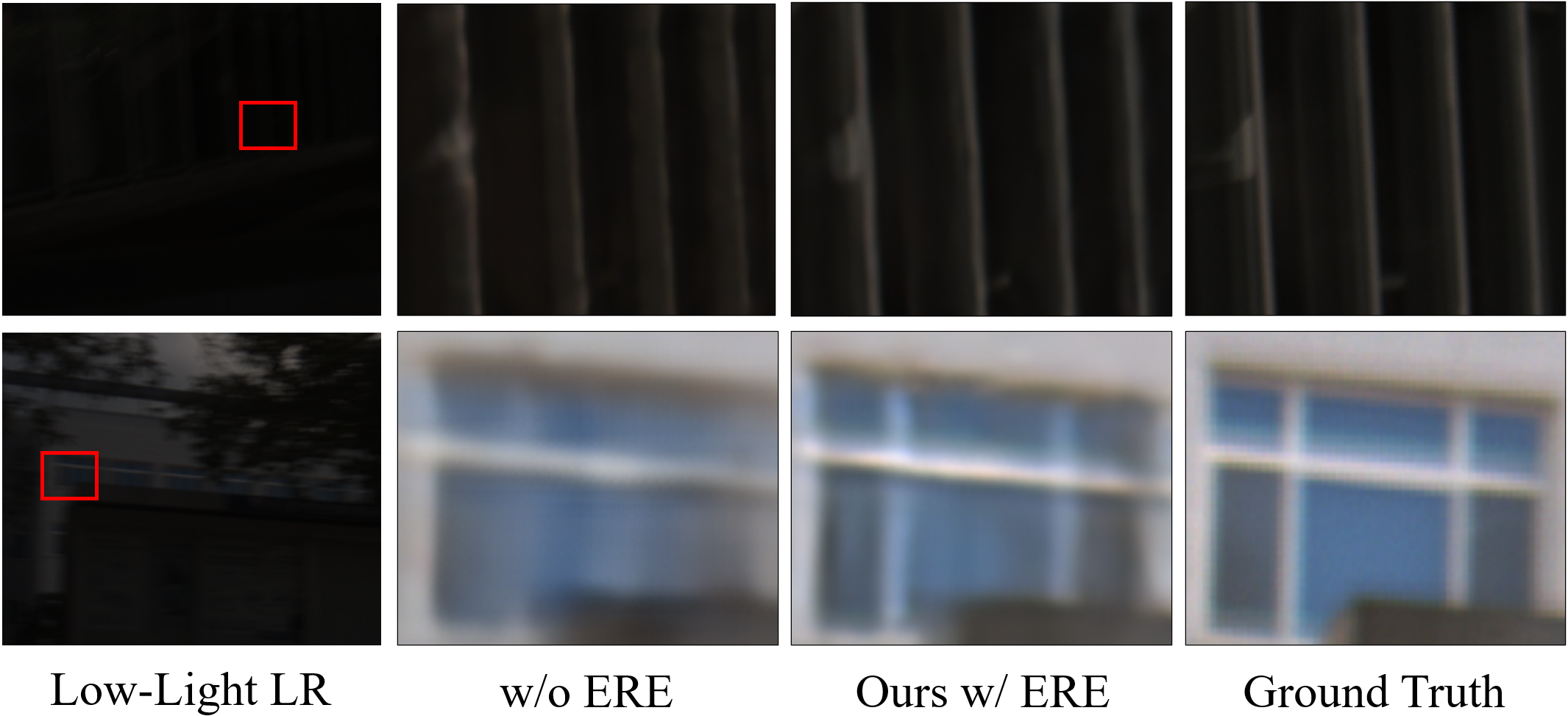}
        \caption{Ablation study of ERE. The model equipped with the ERE module can recover visually clearer results.}
        \label{fig:fig8}
        \vspace{-1ex}
    \end{figure}

    \noindent\textbf{Implementation Details.}\hspace{1mm} Our model is trained from scratch on each dataset. During training, we use 15 input frames with a mini-batch size of 8 and apply center-cropping to both input frames and event voxels to a size of $64 \times 64$. Data augmentation is performed with random horizontal and vertical flips. The model is trained for 300K iterations using the Adam optimizer and Cosine Annealing learning rate scheduler. We adopt the Charbonnier loss~\cite{lai2017deep} for supervision and use SpyNet~\cite{ranjan2017optical} to compute optical flow. We use SCI~\cite{ma2022toward} as our Retinex decomposition model. For SpyNet and SCI, the initial learning rate is $2.5 \times 10^{-5}$, frozen for the first 5K iterations. The initial learning rate for other modules is $2 \times 10^{-4}$. Training is conducted on 2 NVIDIA RTX4090 GPUs, taking about four days per dataset to converge.

    \subsection{Comparisons with State-of-the-Art Methods}

    \noindent\textbf{Baselines.}\hspace{1mm} We compare our method with both RGB-based and event-based SOTA methods, covering two strategies: cascade and one-stage LVSR. For RGB-based VSR, we include BasicVSR++~\cite{chan2022basicvsr++}, DP3DF~\cite{xu2023deep}, MIA-VSR~\cite{zhou2024video}, IART~\cite{xu2024enhancing}, and FMA-Net~\cite{youk2024fma}. For event-based VSR, we compare with EGVSR~\cite{lu2023learning} and EvTexture~\cite{kai2024evtexture}. We also include RGB-based low-light enhancement methods: Retinexformer~\cite{cai2023retinexformer}, FastLLVE~\cite{li2023fastllve}, CoLIE~\cite{chobola2024fast}, and Zero-IG~\cite{shi2024zero}, as well as event-based methods: EvLowLight~\cite{liang2023coherent} and EvLight~\cite{liang2024towards}. As shown in Tab.~\ref{tab:table1}, these baselines are grouped into six categories: (I) RGB-based enhancement $+$ VSR, (II) RGB-based VSR $+$ enhancement, (III) Event-based enhancement $+$ VSR, (IV) Event-based VSR $+$ enhancement, (V) RGB-based joint enhancement and VSR, and (VI) Event-based joint enhancement and VSR. For category (IV), due to the lack of HR events after the first stage, we use CoLIE and Zero-IG as substitutes. Note that all methods are retained on the same dataset for fair comparison.

    \begin{figure}[t!]
        \centering
        \includegraphics[width=\columnwidth]{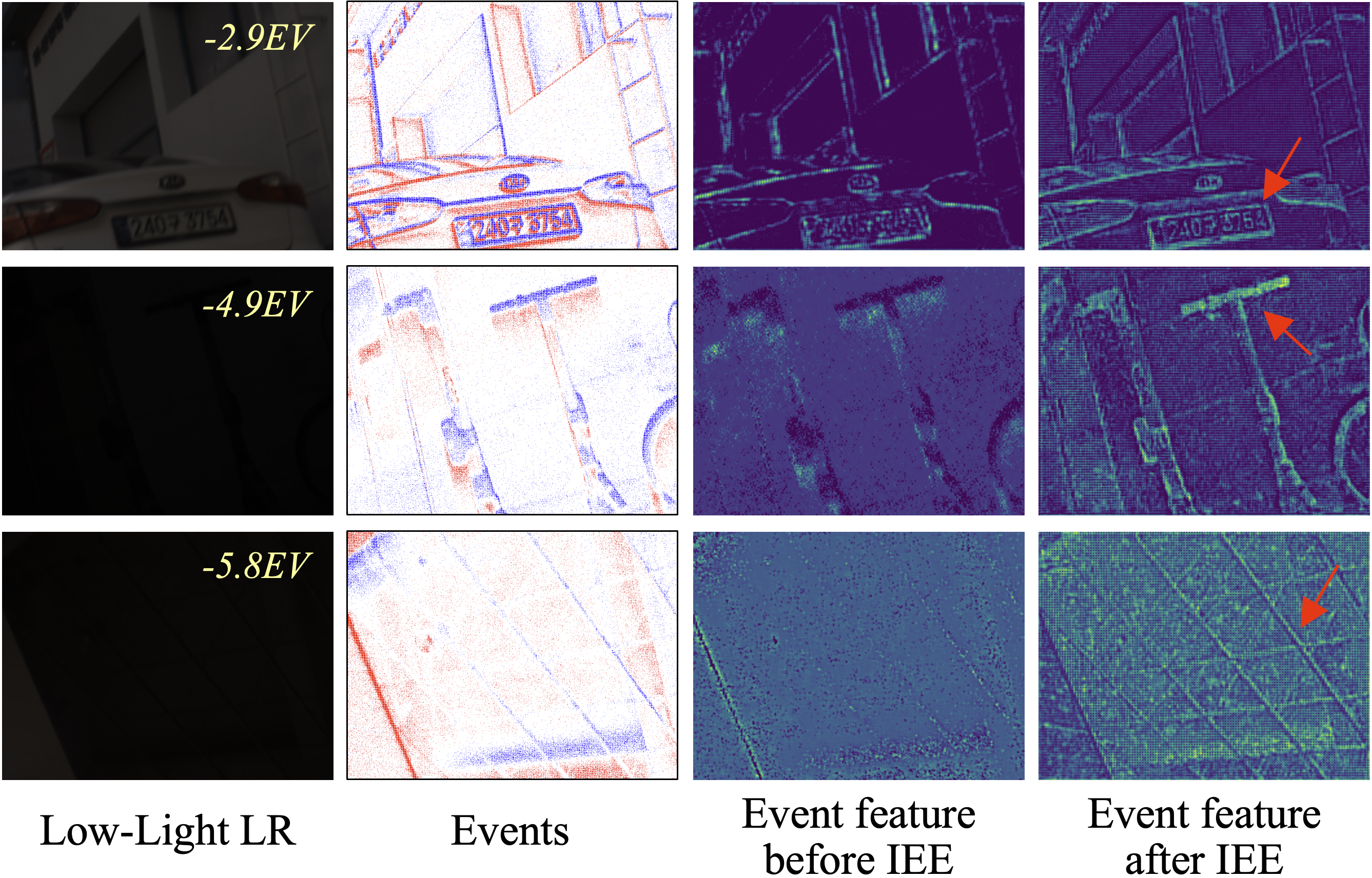}
        \caption{Event features before and after IEE. The module effectively enhances details and suppresses trailing effects.}
        \label{fig:fig9}
        \vspace{-2ex}
    \end{figure}

    \noindent\textbf{Quantitative Results.}\hspace{1mm} Tabs.~\ref{tab:table1} and~\ref{tab:table2} report comparisons in spatial quality (PSNR, SSIM, LPIPS), temporal consistency (tOF~\cite{chu2020learning}, TCC~\cite{chi2020all}), and computational cost. Our RetinexEVSR consistently outperforms all baselines across all datasets. Compared to EvTexture, it improves PSNR by \textbf{2.95}, 0.95, 0.95, 0.93, and 0.85 dB on five datasets, while reducing FLOPs by \textbf{86.1\%} and runtime by 64.9\%, using fewer parameters.

    \noindent\textbf{Qualitative Results.}\hspace{1mm} We also perform qualitative comparisons on these datasets. The visual results are shown in Figs.~\ref{fig:fig5} and~\ref{fig:fig6}. It is obvious that our method enhances illumination to a well-lit level and restores textural details more accurately, while suppressing artifacts. Moreover, the histograms of the RGB channels in each figure show that our method produces color distributions more closely matching those of the ground-truth images.

    \noindent\textbf{Generalization to Real-world SR.}\hspace{1mm} Following prior VSR studies~\cite{kai2024evtexture}, our main experiments use bicubic degradation. To assess real-world generalization, we test the SDE-trained model on SDE without downsampling.  NIQE~\cite{zhang2015feature}, PI~\cite{blau20182018}, CLIP-IQA~\cite{wang2023exploring}, and Q-Align~\cite{wu2024q} are used for no-reference evaluation. As shown in Tab.~\ref{tab:table3}, our method achieves SOTA results on most metrics, especially on SDE-outdoor, highlighting its strong generalization to real-world low-light videos.

    \begin{figure}[t!]
        \centering
        \includegraphics[width=\columnwidth]{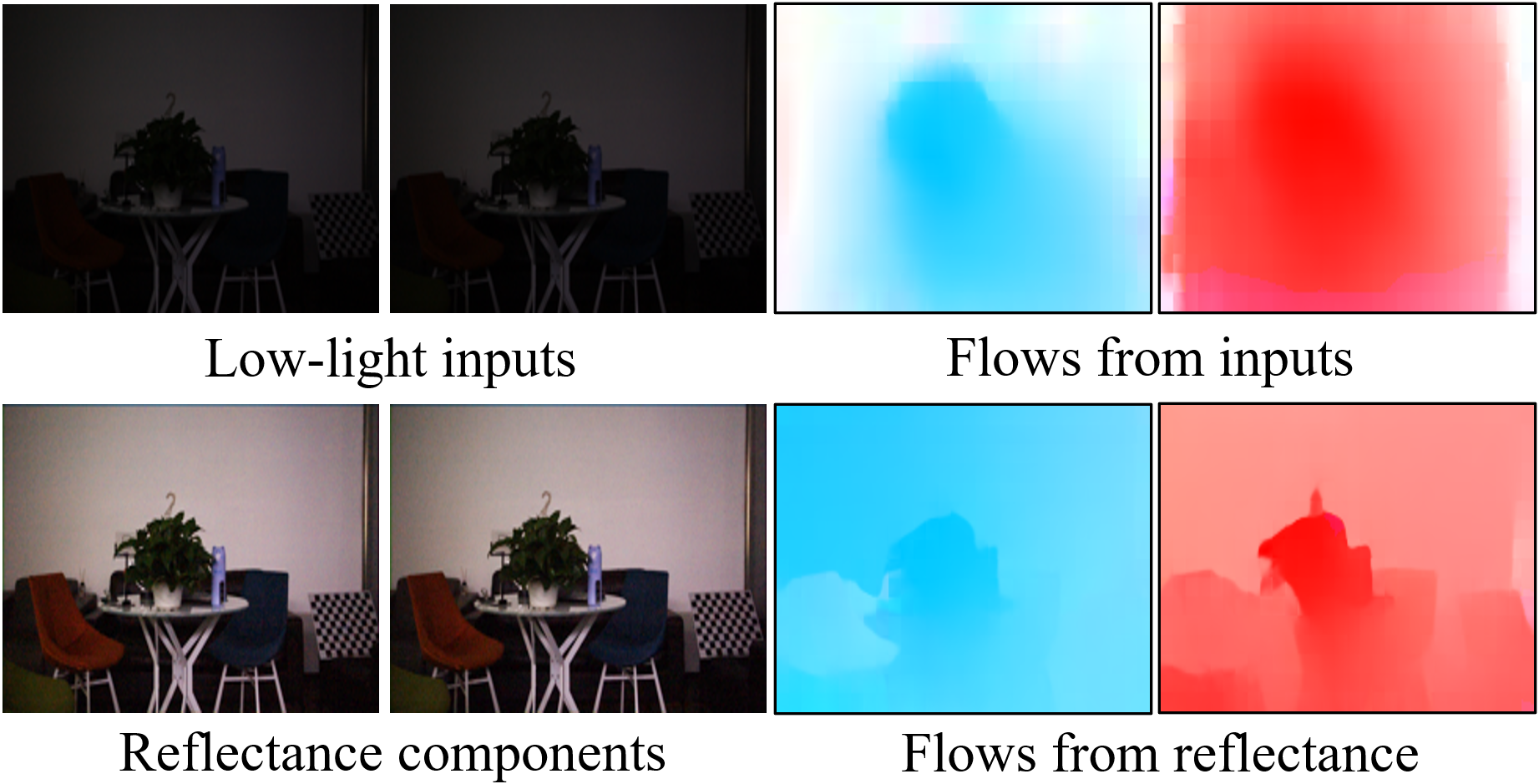}
        \caption{Analysis of optical flow calculation methods. Reflectance-based flow yields sharper edges in low light.}
        \label{fig:fig10}
        \vspace{-1ex}
    \end{figure}

    \subsection{Ablation Study}

    We conduct comprehensive ablation studies on the SDSD-indoor dataset due to its good convergence and stable performance. The ablation results are summarized in Tab.~\ref{tab:table4}.
    
    \noindent\textbf{Break-down Ablations.}\hspace{1mm} Tab.~\ref{tab:table4}(a-c) shows the effect of the IEE and ERE modules. The full model achieves the best performance with only a moderate increase in parameters.
    
    \noindent\textbf{The IEE Module.}\hspace{1mm} Tab.~\ref{tab:table4}(e-f) analyzes the effect of scale in IEE. Using illumination to guide and extract event features from three scales gives the best performance, outperforming the single-scale setup by 1.64 dB. As shown in Fig.~\ref{fig:fig7}, the IEE module helps recover sharper structures and finer details. Fig.~\ref{fig:fig9} further verifies that the IEE module effectively enhances event features by reducing trailing effects.

    \noindent\textbf{The ERE Module.}\hspace{1mm} Tab.~\ref{tab:table4}(g-h) examines the impact of using single-scale and unfused features in ERE. The results show that multi-scale fusion in our full model yields notable gains. As illustrated in Fig.~\ref{fig:fig8}, the full model equipped with the ERE module restores clearer and sharper details.

    \noindent\textbf{Retinex Model.}\hspace{1mm} Tab.~\ref{tab:table4}(i-j) compares different Retinex models. Although URetinex-Net~\cite{wu2022uretinex} has more parameters, its supervised nature limits generalization. Our model with the unsupervised SCI~\cite{ma2022toward,ma2025learning} achieves better results, improving PSNR by 0.66 dB.

    \noindent\textbf{Optical Flow.}\hspace{1mm} Tab.~\ref{tab:table4}(k-l) compares optical flow computed from either low-light frames ${X_t}$ or their reflectance ${R_t}$. Using reflectance improves PSNR by 0.43 dB, thanks to clearer structures that enhance edge localization. Fig.\ref{fig:fig10} shows that flow from reflectance captures sharper edges, making it more reliable for alignment in low-light scenes.

    \section{Conclusion}
    In this paper, we present RetinexEVSR, the first event-driven framework for LVSR. Our method leverages Retinex-inspired priors, coupled with a novel RBF strategy, to effectively fuse degraded RGB and event signals under low-light conditions. Specifically, it includes an IEE module that treats the illumination component, decomposed from the input frames, as a global lighting prior to enhance event features. The refined events are then utilized in the ERE module to enhance reflectance details by injecting high-frequency information. Extensive experiments demonstrate that RetinexEVSR achieves state-of-the-art performance on three datasets, including both synthetic and real-world datasets, and generalizes well to unseen degradations, highlighting its potential for low-light video applications.

    \section*{Acknowledgments}

    We acknowledge funding from the National Natural Science Foundation of China under Grants 62472399 and 62021001.

    \appendix
    \section*{Appendix}

    \section{Cross-modal Fusion Block}

\begin{figure}[h!]
        \vspace{0.1em}
    	\centering
    	\includegraphics[width=\columnwidth]{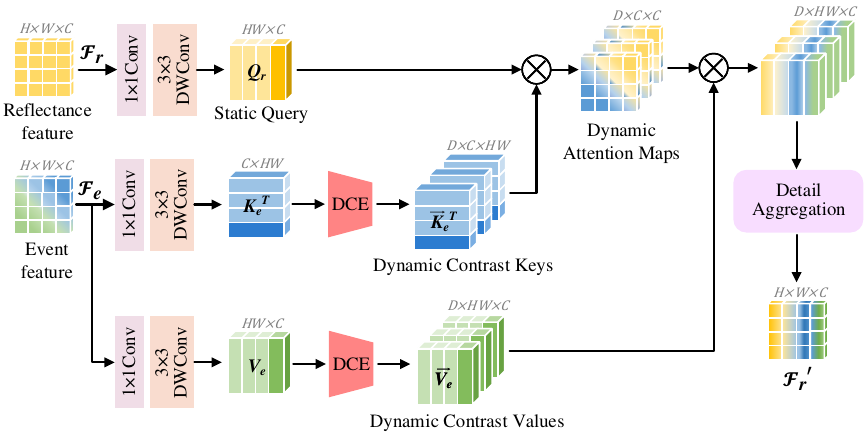}
            \caption{The architecture of the dynamic cross-modal fusion block in ERE. DCE: Dynamic Contrast Extractor.}
    	\label{fig:fig11}
        \vspace{0.1em}
    \end{figure}

    In the context of event-guided reflectance enhancement, effectively fusing dynamic contrast information from events with reflectance features is crucial. Fig.~\ref{fig:fig11} illustrates the event-reflectance dynamic fusion block utilized in our ERE module.

    Unlike standard self-attention mechanisms, our cross-modal attention dynamically integrates information across the event and reflectance modalities. The query $\boldsymbol{Q}_r$ is derived from the reflectance feature map, while the key $\boldsymbol{K}_e$ and value $\boldsymbol{V}_e$ are generated from the event signals. Specifically, given the input features $\mathcal{F}_r, \mathcal{F}_e \in \mathbb{R}^{H \times W \times C}$ from the reflectance and event branches, respectively, the query, key, and value are computed as:
    \begin{equation}
    \boldsymbol{Q}_r = \mathcal{F}_r \boldsymbol{W}^{\mathrm{Q}}, \quad \boldsymbol{K}_e = \mathcal{F}_e \boldsymbol{W}^{\mathrm{K}}, \quad \boldsymbol{V}_e = \mathcal{F}_e \boldsymbol{W}^{\mathrm{V}},\label{eq3}
    \end{equation}
    where $\boldsymbol{W}^{\mathrm{Q}}$, $\boldsymbol{W}^{\mathrm{K}}$, and $\boldsymbol{W}^{\mathrm{V}}$ represent learnable projection matrices, implemented as a stack of $1 \times 1$ and $3 \times 3$ depth-wise convolutions. While vanilla attention computes query-key correlations individually—resulting in a static attention map—it often fails to capture the intricate temporal dynamics of event features $\mathcal{F}_e$. To address this limitation, we propose a dynamic attention mechanism formulation:
    
    \begin{equation}
    \mathcal{F}_r^{'} = \sum_{d=1}^D \alpha\left(\boldsymbol{Q}_r \vec{\boldsymbol{K}}_e^{d^\top}\right) \vec{\boldsymbol{V}}_e^d.\label{eq4}
    \end{equation}
    
    Here, the dynamic keys and values, denoted as $\vec{\boldsymbol{K}}_e$ and $\vec{\boldsymbol{V}}_e \in \mathbb{R}^{D \times C \times N}$, are computed via a Dynamic Contrast Extractor (DCE) employing depth-wise convolutions. The parameter $D=4$ represents the projected temporal dimension within our model. This design empowers the attention mechanism to adaptively capture temporal variations inherent in the event data. The output of the cross-modal attention is computed as shown in Eq.~\ref{eq4}, where $\mathcal{F}_r^{'} \in \mathbb{R}^{H \times W \times C}$ denotes the enhanced reflectance feature, and $\alpha(\cdot)$ represents the softmax function. The symbol $\top$ denotes the matrix transpose operation.
    
    By adaptively focusing on the temporal dynamics of event data, this approach significantly enhances the fusion of event and reflectance features. By leveraging both spatial and temporal contexts, the ERE module improves the robustness and fidelity of reflectance reconstruction, particularly under challenging low-light conditions.

    \begin{figure}[t!]
    	\centering
    	\includegraphics[width=\columnwidth]{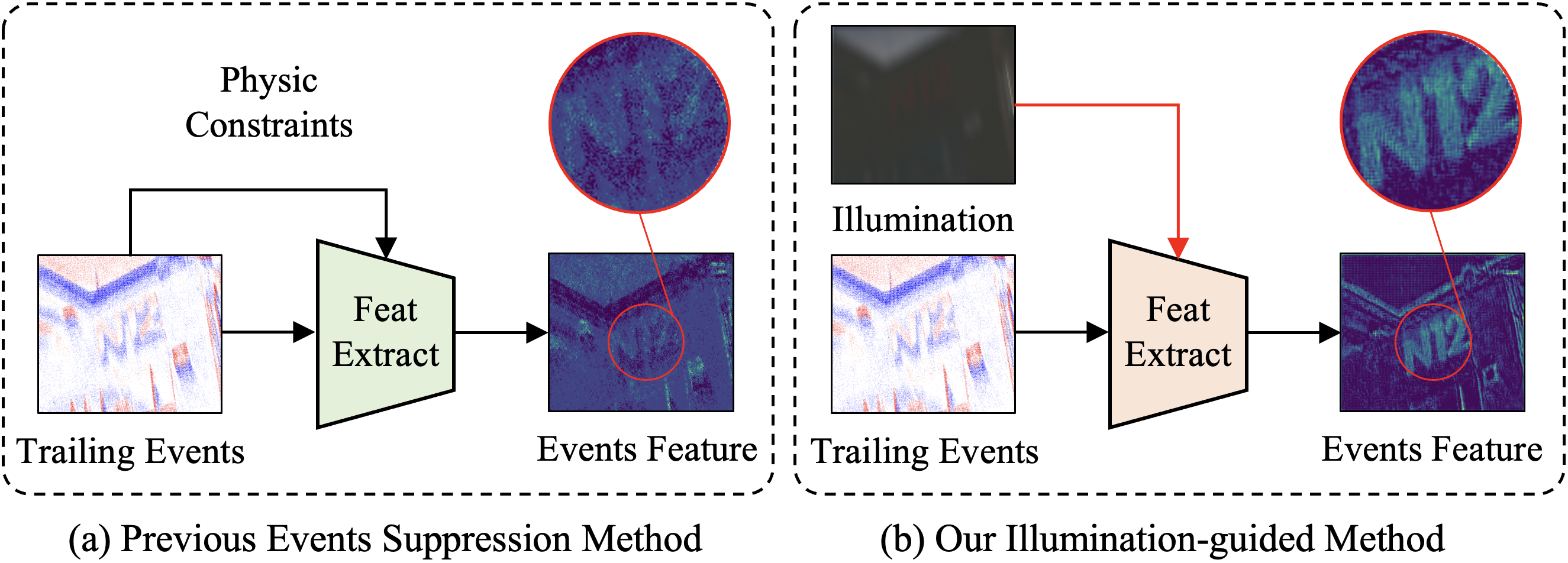}
        \caption{Comparison of event trailing suppression methods. (a) The approach in~\cite{liu2024seeing,liu2025ner} relies solely on the event modality and its physical properties, which proves inadequate for mitigating trailing artifacts. (b) Our illumination-guided method incorporates illumination as a global prior, yielding clearer and sharper event features.}
    	\label{fig:fig12}
    \end{figure}

    \section{Event Suppression Comparison}\label{suppression}
    Mitigating trailing effects is paramount when leveraging event data for low-light VSR, particularly in scenes containing fast-moving objects. Failure to effectively handle these trailing artifacts can blur critical details and degrade the quality of the final reconstruction. Similar challenges have been documented in prior works~\cite{liu2024seeing,liu2025ner}.
    
    In Fig.~\ref{fig:fig12}, we compare the trailing suppression strategy used in these previous methods with our proposed illumination-guided approach. While existing methods rely exclusively on the intrinsic properties of event data, our method introduces illumination information as a global lighting prior to guide the enhancement process. Consequently, our approach more effectively suppresses trailing artifacts induced by low-light conditions, producing more distinct and sharper event features.
    
    \section{More Visual Results}
    
\begin{figure*}[t!]
	\centering
	\includegraphics[width=\textwidth]{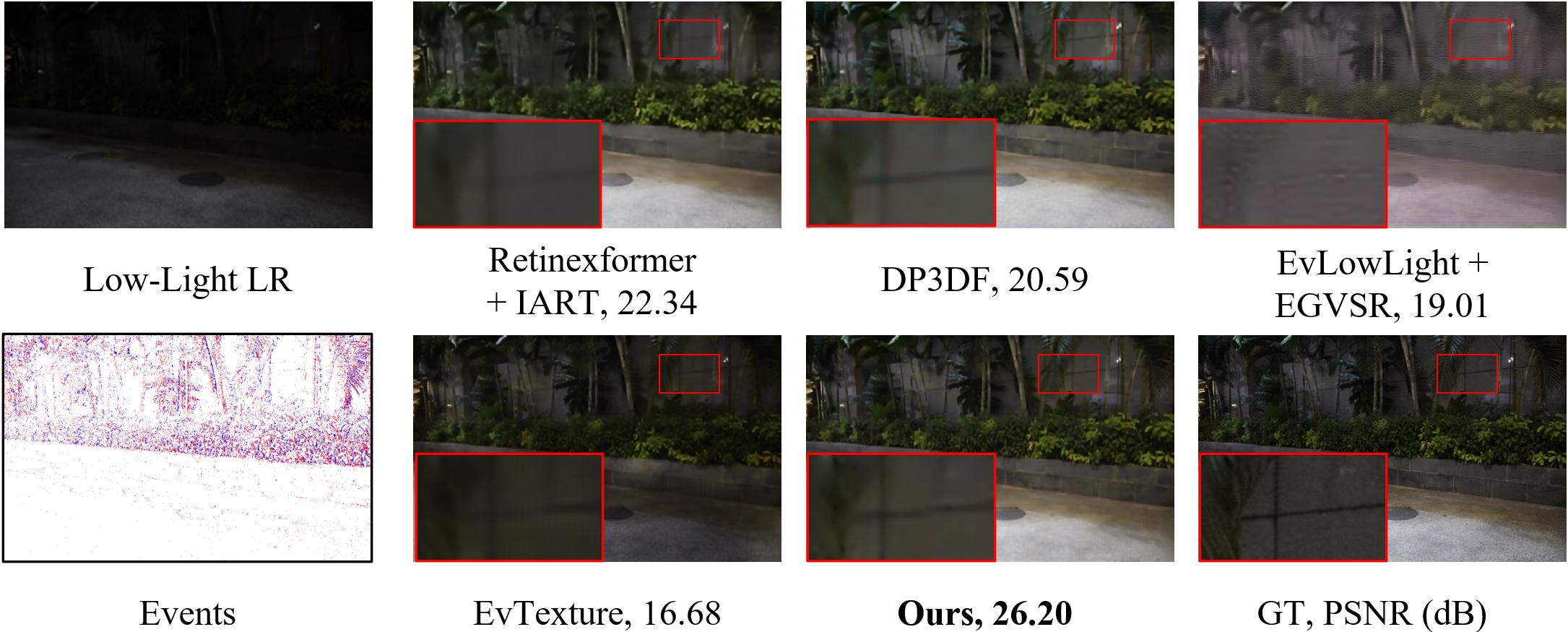}
	\caption{Qualitative comparison on SDSD~\cite{wang2021seeing} for 4$\times$ LVSR. \textbf{Zoomed in for best view.}}
	\label{fig:fig13}
\end{figure*}

\begin{figure*}[t!]
	\centering
	\includegraphics[width=\textwidth]{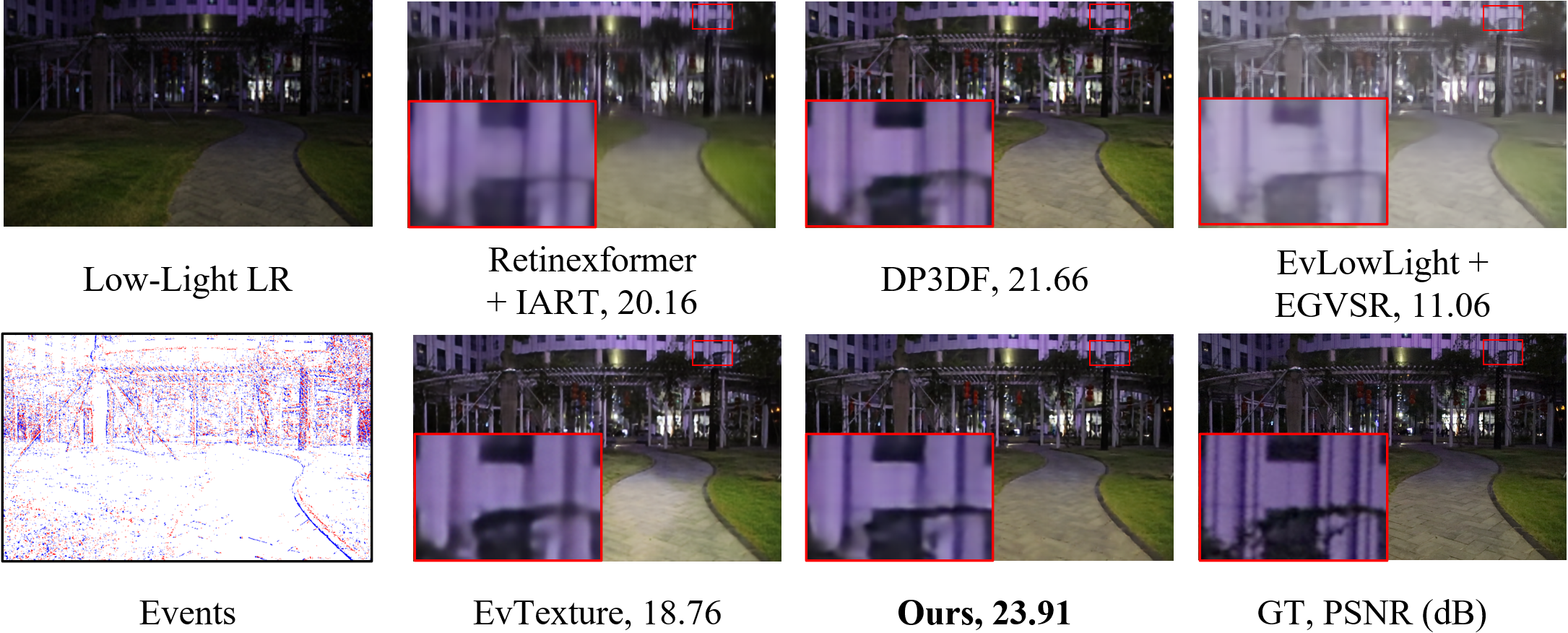}
	\caption{Qualitative comparison on SDSD~\cite{wang2021seeing} for 4$\times$ LVSR. \textbf{Zoomed in for best view.}}
	\label{fig:fig14}
\end{figure*}

\begin{figure*}[t!]
	\centering
	\includegraphics[width=\textwidth]{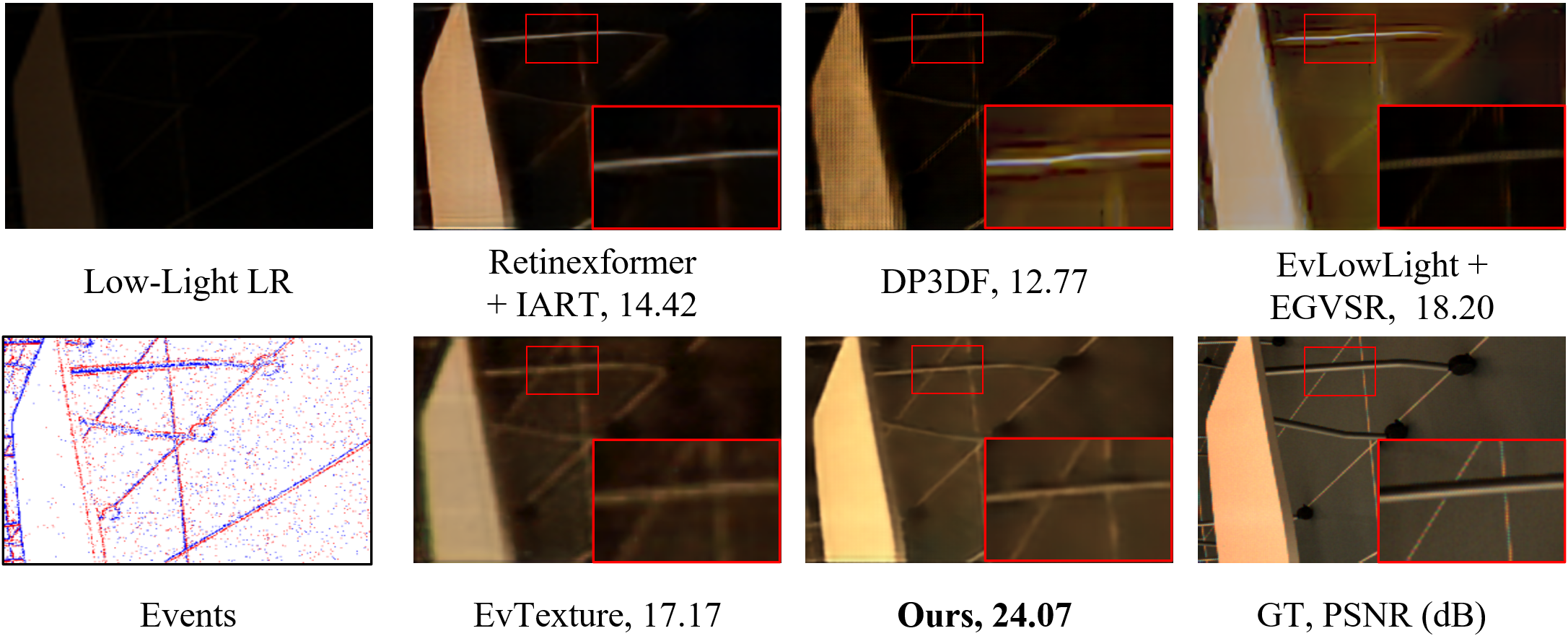}
	\caption{Qualitative comparison on SDE~\cite{liang2024towards} for 4$\times$ LVSR. \textbf{Zoomed in for best view.}}
	\label{fig:fig15}
\end{figure*}

\begin{figure*}[t!]
	\centering
	\includegraphics[width=\textwidth]{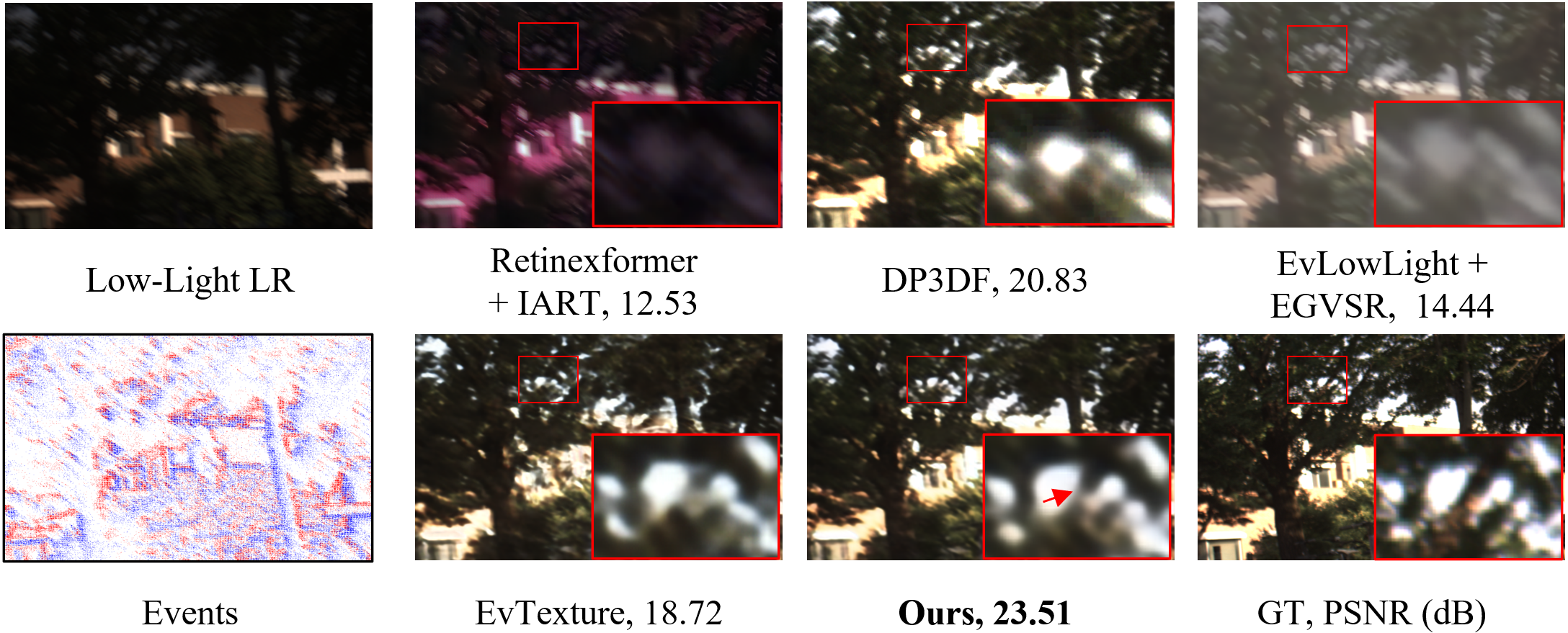}
	\caption{Qualitative comparison on RELED~\cite{kim2024towards} for 4$\times$ LVSR. \textbf{Zoomed in for best view.}}
	\label{fig:fig16}
\end{figure*}

    To further validate the performance of RetinexEVSR, we provide additional visual comparisons on the SDSD \cite{wang2021seeing}, SDE \cite{liang2024towards}, and RELED \cite{kim2024towards} datasets. The results, presented in Figs.~\ref{fig:fig13} through~\ref{fig:fig16}, demonstrate that RetinexEVSR successfully restores complex scenes under low-light conditions, effectively enhancing the visibility of fine details and textures. These compelling results underscore the framework's potential for real-world applications in computational photography and surveillance systems, where low-light video enhancement is critical.

\bibliography{aaai2026}


\clearpage
\newpage

\clearpage
\end{document}